\documentclass[10pt,twocolumn,letterpaper]{article}

\usepackage{cvpr}              
\usepackage{graphicx}
\usepackage{caption}

\usepackage[dvipsnames]{xcolor}

\definecolor{cvprblue}{rgb}{0.21,0.49,0.74}
\usepackage[pagebackref,breaklinks,colorlinks,citecolor=cvprblue]{hyperref}

\newcommand{\parahead}[1]{\noindent\textbf{#1}.\ }

\newcommand{\name}{ActAnywhere\xspace}

\newcommand{\bx}{\mathbf{x}}

\newcommand{\bF}{\mathbf{F}}

\newcommand{\bmu}{\mbox{\boldmath$\mu$}}

\newcommand{\bSigma}{\mbox{\boldmath$\Sigma$}}

\newcommand{\mM}{\mathcal{M}}

\newcommand{\mI}{\mathcal{I}}

\newcommand{\mS}{\mathcal{S}}

\newcommand{\mN}{\mathcal{N}}
\newcommand{\mZ}{\mathcal{Z}}

\def\confName{CVPR}
\def\confYear{2024}

\title{ActAnywhere: Subject-Aware Video Background Generation}

\author{Boxiao Pan$^{1,2\footnote[1]{}}$\qquad Zhan Xu$^{2}$\qquad Chun-Hao Paul Huang$^{2}$ \qquad Krishna Kumar Singh$^{2}$ \\
Yang Zhou$^{2}$ \qquad Leonidas J. Guibas$^{1}$ \qquad Jimei Yang$^{2}$ \\
{$^1$Stanford University \qquad $^2$Adobe Research}
}
\begin{document}
\maketitle

\footnotetext[0]{* Work done during an internship at Adobe.}
\begin{abstract}
Generating video background that tailors to foreground subject motion is an important problem for the movie industry and visual effects community. This task involves synthesizing background that aligns with the motion and appearance of the foreground subject, while also complies with the artist's creative intention. We introduce \name, a generative model that automates this process which traditionally requires tedious manual efforts. Our model leverages the power of large-scale video diffusion models, and is specifically tailored for this task. \name takes a sequence of foreground subject segmentation as input and an image that describes the desired scene as condition, to produce a coherent video with realistic foreground-background interactions while adhering to the condition frame. We train our model on a large-scale dataset of human-scene interaction videos. Extensive evaluations demonstrate the superior performance of our model, significantly outperforming baselines. Moreover, we show that \name generalizes to diverse out-of-distribution samples, including non-human subjects. Please visit our project webpage at \url{https://actanywhere.github.io}.
\vspace{-7mm}
\end{abstract}

\section{Introduction}
Compositing an acting video of a subject onto a novel background is central for creative story-telling in filmmaking and VFX. The key requirement is seamlessly integrating the foreground subject with the background in terms of camera motions, interactions, lighting and shadows, so that the composition looks realistic and vivid as if the subject acts physically in the scene.
In movie industry, this process is often conducted by virtual production~\cite{VirtualProduction} that requires artists to first create a 3D scene and then to film the acting video in an LED-walled studio or to render the video in 3D engines. This process is not only tedious and expensive, but most importantly, prevents artists from quickly iterating their ideas. 

\begin{figure*}[t]
    \centering
    \includegraphics[width=\linewidth]{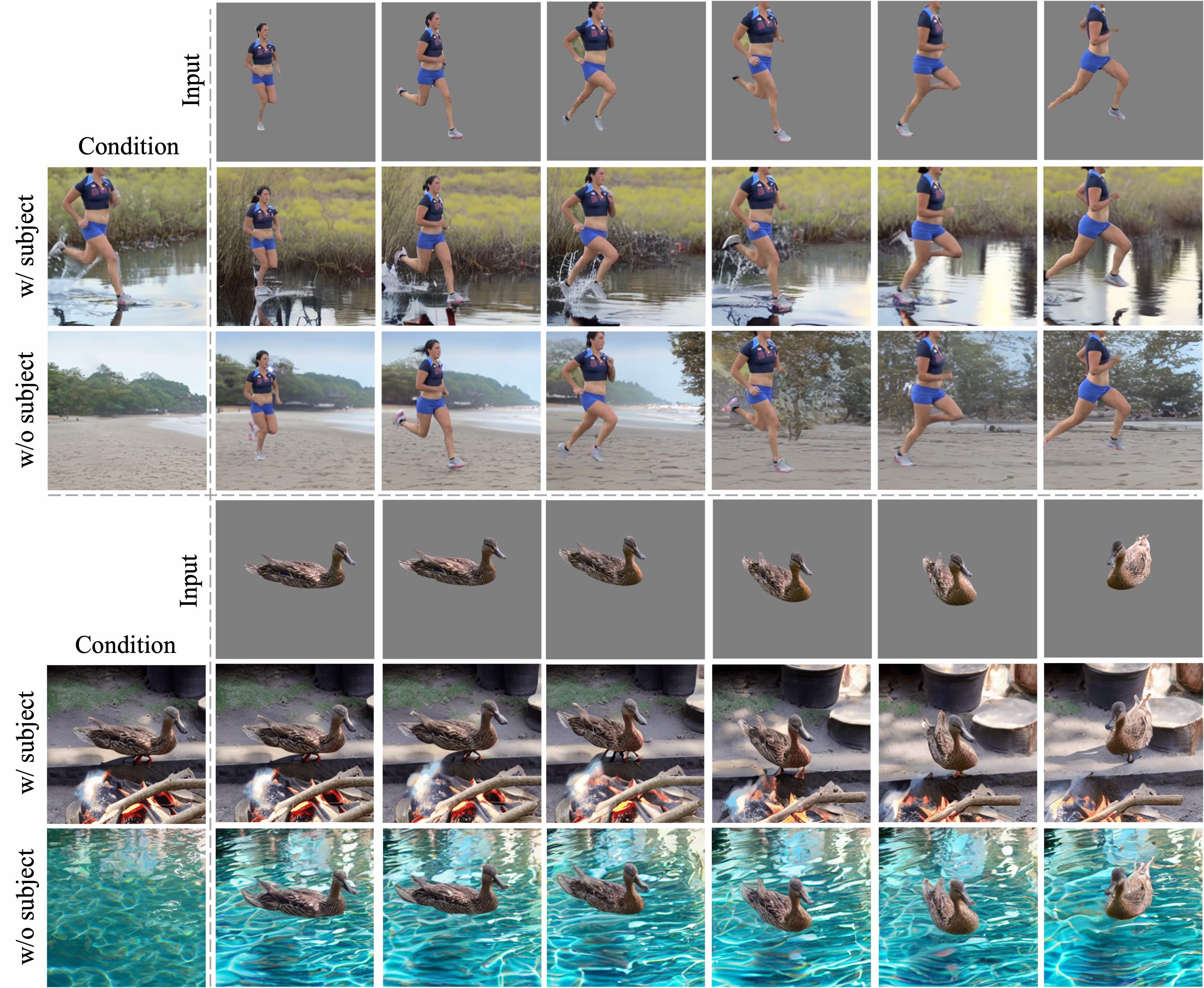}
    \caption{
    Given a sequence of foreground segmentation as input, and one frame that describes the background as the condition, \name generates coherent video background that adapts to the subject motion. We show two subjects here, each with two generated samples. \name is able to generate videos consistent with the condition frame with highly realistic details such as splatting water, moving smoke and flame, shadows, duck feet, etc. It generalizes to a diverse distribution of subjects and backgrounds, including non-human subjects. Our method works with both composited frames and background-only images as the condition.
    }
    \vspace{-3mm}
    \label{fig:teaser}
\end{figure*}

Inspired by this artistic workflow, we study a novel problem of automated subject-aware video background generation. As shown in \cref{fig:teaser}, given a foreground segmentation sequence that provides the subject motion, as well as a condition frame that describes a novel scene, we aim to generate a video that adapts the person to the novel scene with realistically synthesized foreground-background interactions. This condition frame can be either a background-only image, or a composite frame consisting of both  background and foreground, which can be created manually using photo editing tools such as Adobe Photoshop~\cite{adobe_ps} or via automated image outpainting methods such as Dall-E~\cite{DALLE2}.

This problem poses significant challenges, as the human-scene interactions need to be correctly inferred and extrapolated into an extended space-time volume given only the two input signals. The model also needs to implicitly reason about the camera motion from the sequence of foreground segmentation, which is an inherently ambiguous problem. For example, in the first case of \cref{fig:teaser}, the model needs to generate a background that moves according to the direction that the woman runs towards. Last but not the least, to support various applications, we aim to have a model with strong generalization capability, allowing for the realistic and creative integration of different subjects into diverse background scenes.

Existing works on video generation and editing, despite achieving impressive progress, are not able to solve this task. Recent approaches generally focus on unconditional video generation~\cite{TATS2022,LVDM2022,PVDM2023}, text-conditioned video generation~\cite{TATS2022,LVDM2022,Phenaki2022,AnimateDiff2023,AlignLatents2023}, or simple outpainting masked regions~\cite{MAGVIT2023,ProPainter2023}. Meanwhile, video editing methods assume a source video as input and make edits based on some condition signals, most commonly natural language~\cite{Text2LIVE2022,Pix2Video2023,TuneAVideo2023,Gen12023,TokenFlow2023,VideoControlNet2023,T2VZero2023}. However, the edits these method make are mostly limited to stylization, which means they preserve the spatial structure in the source video and perform only stylizing changes. On the other hand, simply propagating image outpainted results~\cite{AnimateDiff2023,VideoCrafter2023} does not necessarily respect the guidance from the foreground subject motion, and hence is under-constrained (as shown later in \cref{subsec:comparison}).
In this paper, we aim to completely generate the structure and texture of the video background, while keeping it coherent with the foreground subject motion. 

To this end, we propose a diffusion-based model that leverages cross-frame attention for temporal reasoning. Specifically, our model takes as input a sequence of segmented foreground subject, the corresponding masks, and a single condition frame of the background, to generate the composited video with a hallucinated video background. Since temporal attention is currently the standard de facto for diffusion-based video generation~\cite{AnimateDiff2023,TuneAVideo2023,ControlAVideo2023,Gen12023,TokenFlow2023,T2VZero2023} due to the ability to generate temporally coherent videos, we also perform temporal self-attention to frame-wise features, while conditioning the diffusion process on the features of the background frame.

We train our model on a large-scale dataset~\cite{PutPeople2023} that consists of 2.4M videos of human-scene interactions in a self-supervised fashion, and evaluate both on a held-out set as well as on videos from DAVIS~\cite{DAVIS2017}. \name is able to generate highly realistic videos that follow the condition frame, and at the same time synthesizes video background that conforms to the foreground motion. Notably, despite trained solely on videos of humans, \name generalizes to non-human subjects in a zero-shot manner.

In summary, our contributions are:
\begin{enumerate}
    \item We introduce a novel problem of automated subject-aware video background generation. 
    \item We propose \name, a video diffusion-based model to solve this task, and train it on a large-scale human-scene interaction video datatset in a self-supervised manner.
    \item Extensive evaluations demonstrate that our model generates coherent videos with realistic subject-scene interactions, camera motions, lighting and shadows, and generalizes to out-of-distribution data including non-human subjects, such as animals and man-made objects.
\end{enumerate}

\section{Related Work}
\parahead{Video generation}
There have been a long thread of works on video generation. The core architecture has evolved from  GANs~\cite{SceneDynam2016,MOCOGAN2018,GANVideoGen2019} to  more recent transformers~\cite{MAGVIT2023,TATS2022,Phenaki2022,ProPainter2023} and diffusion models~\cite{LVDM2022,ImagenVideo2022,PVDM2023,T2VZero2023,AlignLatents2023,VideoCrafter2023,AnimateDiff2023}. Below we review most related diffusion-based works. Most of these works leverage temporal self-attention blocks inside the denoising U-Net in order to acquire temporal awareness. On top of that, Text2Video-Zero~\cite{T2VZero2023} introduces additional noisy scheduling to correlate the latents in a video. LVDM~\cite{LVDM2022} and Align Your Latents~\cite{AlignLatents2023} both design a hierarchical approach to generate longer-term videos. Align Your Latents additionally fine-tunes a spatial super-resolution model for high-resolution video generation. AnimateDiff~\cite{AnimateDiff2023} proposes to train the temporal attention blocks on a large-scale video dataset, which can then be inserted into any text-to-image diffusion models (given that the architecture fits) to turn that into a text-to-video model, in a zero-shot manner. VideoCrafter1~\cite{VideoCrafter2023} further uses dual attention to enable joint text and image-conditioned generation. 
These works focus on unconditional generation or with text or image conditioning, but are not able to follow the guidance of additional foreground motion. 

\parahead{Video editing}\noindent
Another thread studies the problem of video editing, where a source video is given as input, and edits are performed according to some condition signals. Text2Live~\cite{Text2LIVE2022} uses pre-trained video atlases of the input video, and performs text-guided edits on the foreground or background. Gen1~\cite{Gen12023} leverages depth maps estimated by a pre-trained network~\cite{MiDaS2020} as an additional condition to improve the structure consistency. Tune-A-Video~\cite{TuneAVideo2023} proposes to finetune only part of the spatial-attention blocks and all of the temporal-attention blocks on a single input video. TokenFlow~\cite{TokenFlow2023} uses latent nearest neighbor fields computed from the input video to propagate edited features across all frames. Both VideoControlNet~\cite{VideoControlNet2023} and Control-A-Video~\cite{ControlAVideo2023} adopt a ControlNet-like approach~\cite{ControlNet2023} to condition the video diffusion process with additional signals such as depth maps or Canny edges extracted from the input video. One downside of these works is that the generated videos tend to keep the spatial structure from the source video, which greatly limits the edits that the model is able to perform. In our work, we propose to condition on the foreground segmentation for the motion, while extract the background information only from one condition frame. In particular, using the masked foreground as input endows a nice separation as in what to preserve and what to generate.

\parahead{Image and video inpainting}
Image / video inpainting aims to fill a missing region, often expressed as a mask. These methods either take condition signals such as natural language and image~\cite{StableDiffusion2022,SmartBrush2023,PaintbyExample2023}, or rely solely on the context outside the masked region~\cite{Palette2022,MAGVIT2023,M3DDM2023,ProPainter2023}. Recent diffusion-based image inpainting methods use a combination of masked image and the mask itself, and condition the diffusion process either on natural language~\cite{StableDiffusion2022,SmartBrush2023} or an image of the condition object~\cite{PaintbyExample2023}, or perform unconditional diffusion~\cite{Palette2022}. For video in-painting, MAGVIT~\cite{MAGVIT2023} proposes a generative video transformer trained through masked token prediction, and is able to inpaint small masked regions afterwards. ProPainter~\cite{ProPainter2023} designs a flow-based method by propagating pixels and features through completed flows. M3DDM~\cite{M3DDM2023} leverages a video diffusion model, and conditions the diffusion process on global video features extracted by a video encoder. Different from these works, we aim to generate large background regions that strictly follow the condition frame. Moreover, the generated background needs to adapt to the foreground subject motion in a coherent way. This poses significant challenges that previous inpainting methods cannot tackle.

\section{Method}
We first provide essential preliminary background on latent diffusion in~\cref{subsec:bg_diffusion}. We then formally define our problem in~\cref{subsec:problem_formulation} and delve into our model design in~\cref{subsec:model}. Finally, we specify the training details in~\cref{subsec:training}.

\begin{figure*}[t]
    \centering
    \includegraphics[width=\linewidth]{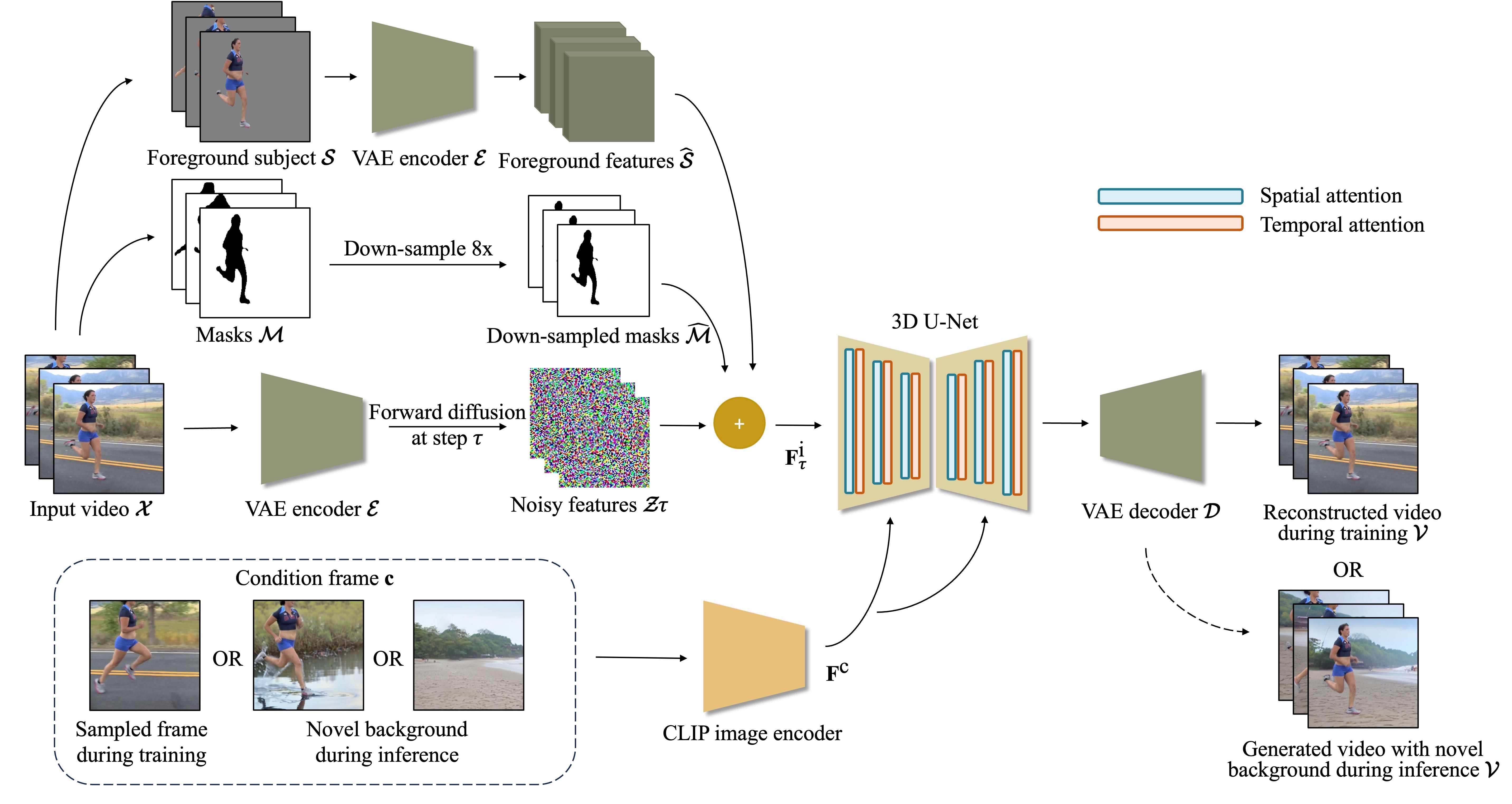}
    \caption{
    \textbf{Architecture overview}. During training, we take a randomly sampled frame from the training video to condition the denoising process. At test time, the condition can be either a composited frame of the subject with a novel background, or a background-only image. 
    }
    \vspace{-4mm}
    \label{fig:pipeline}
\end{figure*}

\subsection{Preliminaries on Latent Diffusion Models}
\label{subsec:bg_diffusion}
Diffusion models such as DDPM~\cite{DDPM2020}, encapsulate a forward process of adding noise and a backward process of denoising. Given a diffusion time step $\tau$, the forward process incrementally introduces Gaussian noises into the data distribution $x_0 \sim q(x_0)$ via a Markov chain, following a predefined variance schedule denoted as $\beta$:
\begin{equation}
    q(\bx_{\tau}|\bx_{{\tau}-1})=\mN(\bx_{\tau};\sqrt{1-\beta_{\tau}}\bx_{{\tau}-1},\beta_{\tau}\mI)
\label{eq:diffusion_forward}
\end{equation}

For the backward process, a U-Net~\cite{U-Net} $\epsilon_{\theta}$ is trained to denoise $\bx_{\tau}$ and recover the original data distribution:
\begin{equation}
    p_{\theta}(\bx_{\tau-1}|\bx_{\tau})=\mN(\bx_{{\tau}-1};\bmu_{\theta}(\bx_{\tau}, \tau),\bSigma_{\theta}(\bx_{\tau},\tau))
\label{eq:diffusion_backward}
\end{equation}
$\bmu_{\theta}$ and $\bSigma_{\theta}$ are parametrized by $\epsilon_{\theta}$. The discrepancy between the predicted noise and the ground-truth noise is minimized as the training objective.

Stable Diffusion~\cite{StableDiffusion2022} further proposes to train the diffusion model in the latent space of a VAE~\cite{kingma2013auto}. Specifically, an encoder $\mathcal{E}$ learns to compress an input image $x$ into latent representations $z = \mathcal{E}(x)$, and a decoder $\mathcal{D}$ learns to reconstruct the latents back to pixel space, such that $x = \mathcal{D}(\mathcal{E}(x))$. In this way, the diffusion is performed in the latent space of the VAE.

\subsection{Problem Formulation}
\label{subsec:problem_formulation}
Given an input video $\mathcal{X} \in \mathbb{R}^{T \times H \times W \times 3}$ featuring a foreground subject, we first deploy a segmentation algorithm, such as Mask R-CNN~\cite{he2017mask}, to obtain a subject segmentation sequence, $\mathcal{S} \in \mathbb{R}^{T \times H \times W \times 3}$, along with the corresponding masks, $\mathcal{M} \in \mathbb{R}^{T \times H \times W \times 1}$.
Both $\mathcal{S}$ and $\mathcal{M}$ serve as input to our model. $\mathcal{S}$ contains the segmentation of the foreground subject, with background pixels set to 127 (grey). $\mathcal{M}$ has the foreground pixels set to 0 and background to 1.
Across all our experiments, $H = W = 256$ and $T = 16$.

Additionally, we also incorporate a single condition frame $\mathbf{c} \in \mathbb{R}^{H \times W \times 3}$ describing the background that we want to generate. As shown in \cref{fig:pipeline}, $\mathbf{c}$ is a randomly sampled frame from $\mathcal{X}$ at training time, while can be either a frame showing foreground-background composition or a background-only image at inference time. The goal is thus to generate an output video $\mathcal{V}$ with the subject dynamically interacting with the synthesized background.
The motivation of using an image not language as the condition is that image is a more straightforward media to carry detailed and specific information of the intended background, especially when users already have a pre-defined target scene image. 

\subsection{Subject-Aware Latent Video Diffusion}
\label{subsec:model}
We build our model based on latent video diffusion models~\cite{AnimateDiff2023}. In our architecture design, we address two main questions: 1) providing the foreground subject sequence to the network to enable proper motion guidance, and 2) injecting the condition signal from the background frame to make the generated video adhere to the condition.

We present our pipeline as shown in \cref{fig:pipeline}. 
For the foreground segmentation sequence $\mS$, we use the pre-trained VAE~\cite{StableDiffusion2022} encoder $\mathcal{E}$ to encode the foreground segmentation into latent features $\hat{\mS} \in \mathbb{R}^{16 \times 32 \times 32 \times 4}$. 
We downsample the foreground mask sequence $\mM$ 8 times to obtain the resized mask sequence $\hat{\mM} \in \mathbb{R}^{16 \times 32 \times 32 \times 1}$ to align with the latent features $\hat{\mS}$.
To train the denoising network $\epsilon_{\theta}$, we encode the original frames $\mathcal{X}$ with the same VAE encoder into latent representation $\mZ \in \mathbb{R}^{16 \times 32 \times 32 \times 4}$, and add noises at diffusion time step $\tau$ with the forward diffusion process denoted in \cref{eq:diffusion_forward} to get noisy latent feature $\mZ_{\tau}$.  We subsequently concatenate $\hat{\mS}$, $\hat{\mM}$ and $\mZ_{\tau}$ along the feature dimension, forming a 9-channel input feature $\mathbf{F}^{i}_{\tau} \in \mathbb{R}^{16 \times 9 \times 32 \times 32}$ to the U-Net. During inference, $\mZ_0$ is initialized as Gaussian noises, and gets auto-regressively denoised for multiple time steps to sample a final result, according to the backward diffusion process described in \cref{eq:diffusion_backward}. The denoised latents are then decoded to a video via the VAE decoder $\mathcal{D}$.

We build our 3D denoising U-Net based on AnimateDiff~\cite{AnimateDiff2023}. AnimateDiff works by inserting a series of motion modules in between the spatial attention layers in the denoising U-Net of a pre-trained T2I diffusion model. These motion modules consist of a few feature projection layers followed by 1D temporal self-attention blocks. 

For the condition image $\mathbf{c}$, we follow prior works~\cite{PutPeople2023} to encode it with the CLIP image encoder~\cite{CLIP2021}, and take the features from the last hidden layer as its encoding $\mathbf{F}^{c}$. These features are then injected into the UNet $\epsilon_{\theta}$ through its cross-attention layers, similar to \cite{StableDiffusion2022, PutPeople2023}. We empirically find that this method achieves better temporal consistency compared to other alternatives, such as using VAE features for either cross-attention or concatenation with other input features.

\subsection{Training}
\label{subsec:training}
Model training is supervised by a simplified diffusion objective, namely predicting the added noise~\cite{DDPM2020}:
\begin{equation}
    \mathcal{L} = ||\epsilon - \epsilon_{\theta} (\bF_{\tau}^{i}, \tau, \bF^{c})||_2^2
\end{equation}
where $\epsilon$ is the ground-truth noise added.

\parahead{Dataset}
We train on the large-scale dataset compiled and processed by~\cite{PutPeople2023}, which we refer to as HiC+. The resulting dataset contains 2.4M videos of human-scene interactions. It also provides foreground segmentation and masks. We refer the reader to the original paper for more details.

\parahead{Pre-trained weights}
We initialize the weights of our denoising network $\epsilon_{\theta}$ with the pre-trained weights from the Stable Diffusion image inpainting model~\cite{StableDiffusion2022}, which is fine-tuned on top of the original Stable Diffusion on the text-conditioned image inpainting task. We initialize the weights of the inserted motion modules with AnimateDiff v2\footnote{https://github.com/guoyww/animatediff/}.

For the CLIP image encoder, we use the ``clip-vit-large-patch14" variant\footnote{https://huggingface.co/openai/clip-vit-large-patch14} provided by OpenAI, whose features from the last hidden layer have a dimension of 1024, while the pre-trained U-Net takes in features of dimension 768 as the condition, which are also in the text feature space. To account for this, we train an additional two-layer MLP to project the features into the desired space.

During training, we freeze the shared VAE and the CLIP encoder, and fine-tune the U-Net with the motion modules.

\parahead{Data processing and augmentation}
Obtaining perfect segmentation masks from videos is challenging. The masks may be incomplete, missing some parts of the foreground, or be excessive such that they include leaked background near the boundary. To deal with incomplete segmentation, during training, we apply random rectangular cut-outs to the foreground segmentation and masks. To reduce information leak from excessive segmentation, we perform image erosion to the segmentation and masks with a uniform kernel of size 5 $\times$ 5, both during training and inference.

\parahead{Random condition dropping}
In order to enable classifier-free guidance at test time, we randomly drop the segmentation and the mask, the condition frame, or all of them at 10\% probability each during training. In these cases we set them to zeros before passing into the respective encoders.

\parahead{Other details}
We use the AdamW~\cite{AdamW2017} optimizer with a constant learning rate of 3e-5. We train our model on 8 NVIDIA A100-80GB GPUs at a batch size of 4, which takes approximately a week to fully converge.
\section{Experiments}
We start by describing the data used for evaluation. We then show diverse samples generated from our method in~\cref{subsec:qualitative}, both using an inpainted frame and a background-only frame as conditioning. In \cref{subsec:comparison}, we compare with various baselines. We provide additional results and analysis in \cref{subsec:addt_results}. Specifically, we show that certain general video inpainting capability emerges from our model once trained. We also demonstrate that our model is robust to inaccurate foreground segmentation at test time. Finally, we analyze the model runtime. 

Following prior works~\cite{Text2LIVE2022,TuneAVideo2023,Gen12023,ControlAVideo2023,TokenFlow2023}, we compare with previous works on videos sampled from the DAVIS~\cite{DAVIS2017} dataset. We select videos with both human and non-human subjects. We also evaluate on held-out samples from the HiC+ dataset. Samples with our method are generated with 50 denoising steps, with a guidance scale~\cite{StableDiffusion2022} of 5.

\subsection{Diverse Generation with \name}
\label{subsec:qualitative}

\begin{figure*}
    \centering
    \includegraphics[width=\linewidth]{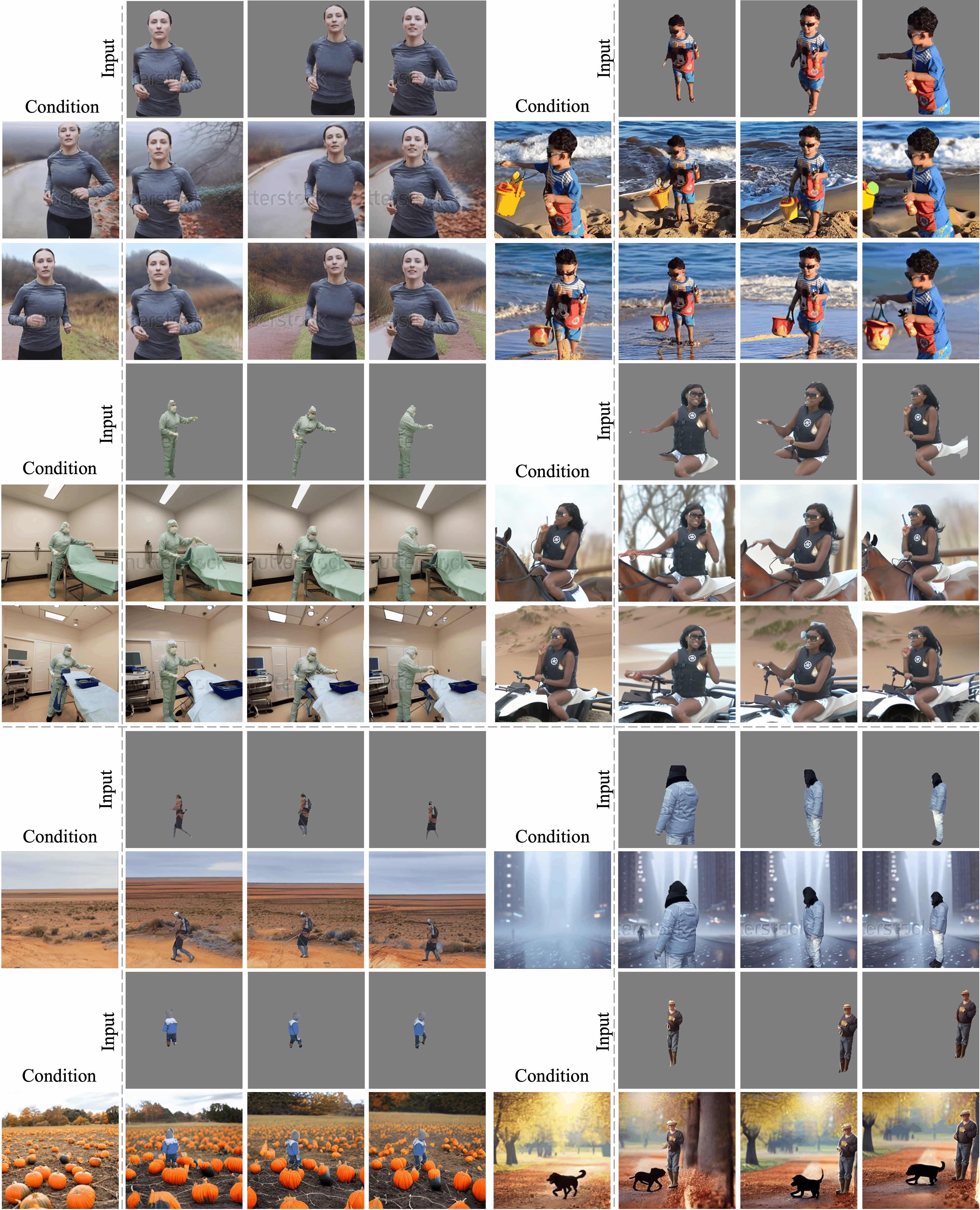}
    \caption{\textbf{Additional results with our method}. The top part shows examples using inpainted frames as condition, while bottom contains examples with background-only conditioning. Foreground sequences are from the held-out set of HiC+.}
    \vspace{-8px}
    \label{fig:addt_qual}
\end{figure*}

\begin{figure*}
    \centering
    \includegraphics[width=\linewidth]{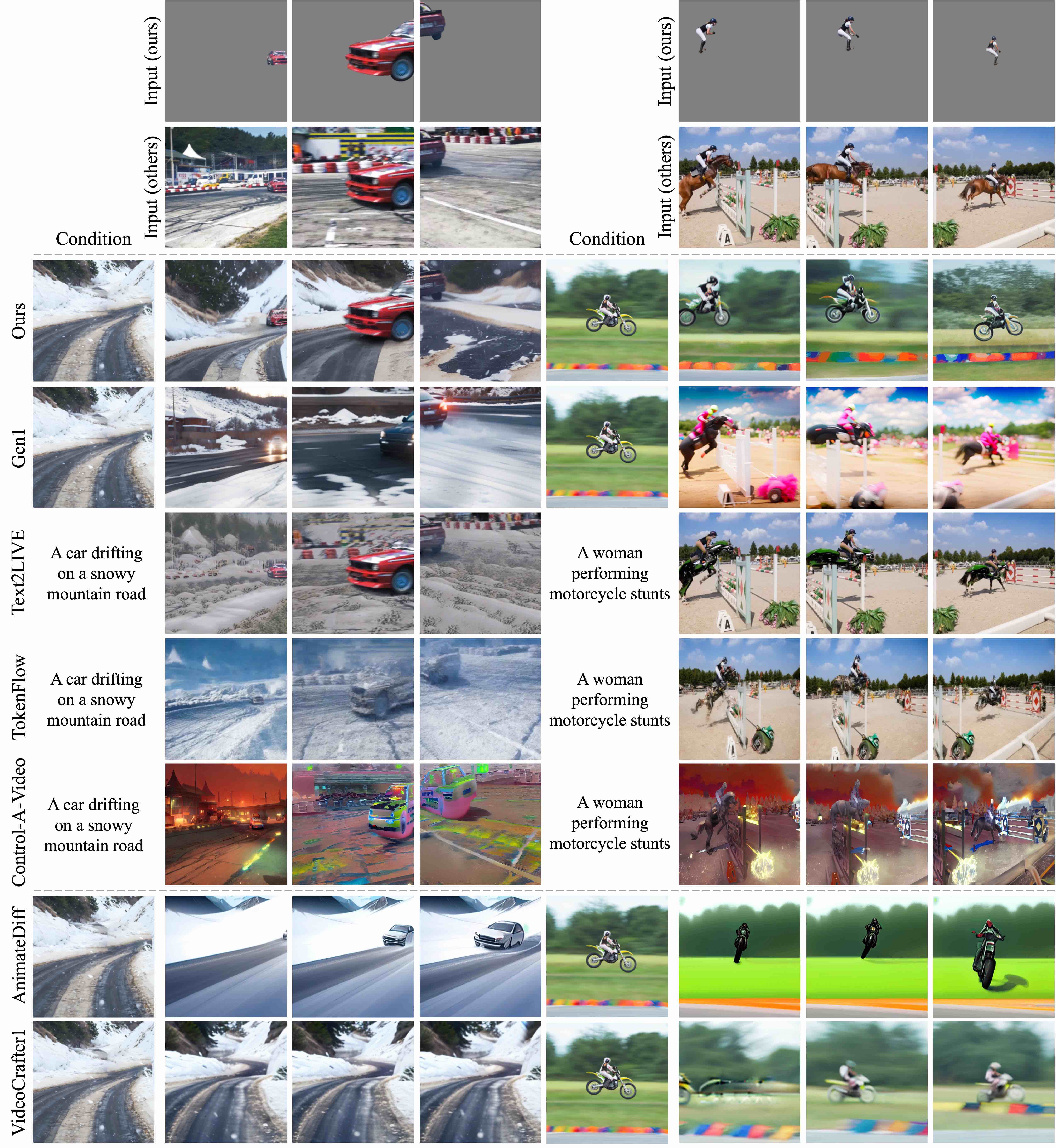}
    \caption{\textbf{Comparison with baselines.} We provide results on two videos sampled from the DAVIS~\cite{DAVIS2017} dataset. For each example, we show three representative frames (top) and their corresponding condition signal (left). Note that different methods assume different input, conditioning or pre-trained models, as specified in \cref{subsec:comparison}.
    }
    \vspace{-4mm}
    \label{fig:exp_comparison}
\end{figure*}

In \cref{fig:addt_qual}, we show results on the held-out segmentation sequences from the HiC+ dataset, using an inpainted frame or a background-only frame as condition. \name generates highly realistic foreground-background interactions both at coarse and fine-grained levels. At a coarse level, our model synthesizes road structure, pumpkin field, city views, waves, etc. that align with the subject's motion. While at a fine-grained level, our method also generates small moving objects that are in close interaction with the subject, such as the buckets, bed sheets, horses and dune buggies, as well as the dog. Moreover, these generation stay consistent across frames, and tightly follow the guidance in the condition frame. The synthesized backgrounds also exhibit coherent scale, lightning, and shadows (also see \cref{fig:teaser}).

\begin{figure*}
    \centering
    \includegraphics[width=\linewidth]{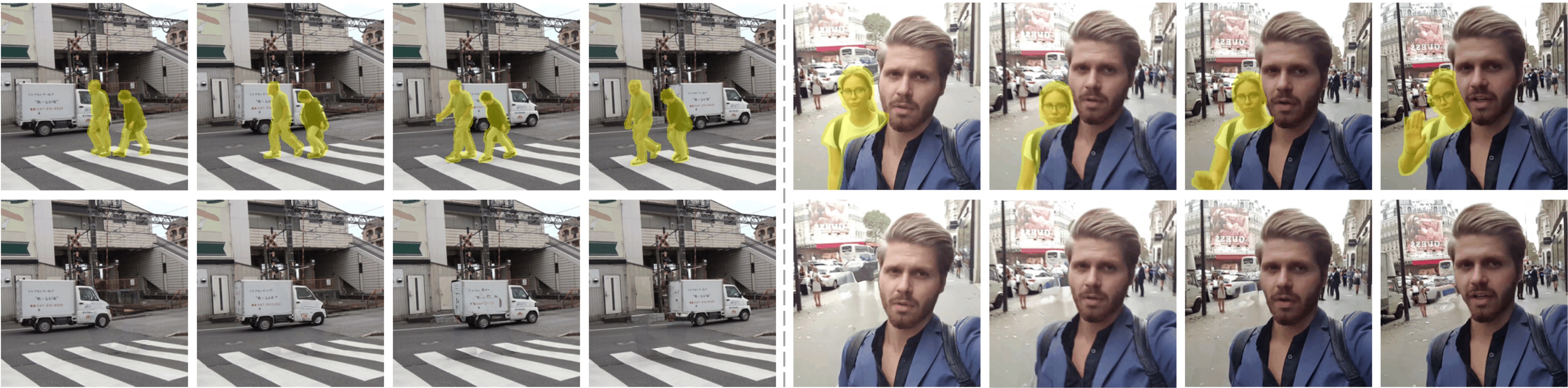}
    \vspace{-16px}
    \caption{\textbf{Zero-shot video inpainting with our model.} We show two cases from DAVIS, each with four sampled frames. The yellow regions denote the masked areas to be inpainted.}
    \vspace{-4mm}
    \label{fig:inpainting_app}
\end{figure*}

\subsection{Comparison with Baselines}
\label{subsec:comparison}

\parahead{Baselines}
We first clarify that since we study a novel problem, there is no prior work operating under the exact same setting to the best of our knowledge. We hence compare to closest works and adapt some, \ie AnimateDiff~\cite{AnimateDiff2023}, if necessary. Nonetheless, we emphasize that the formulation and pipeline are the core contribution of this work.

We compare \name to a number of baselines, which we classify based on whether they do (\cref{fig:exp_comparison} top) or do not (\cref{fig:exp_comparison} bottom) take a video as input. For the methods taking a video as input, Gen1~\cite{Gen12023} takes an additional image as condition, and also leverages a pre-trained depth-estimation network~\cite{MiDaS2020}. Given pre-trained neural atlases~\cite{NLA2021}, Text2LIVE~\cite{Text2LIVE2022} assumes a text prompt as condition to synthesize the edited video. TokenFlow~\cite{TokenFlow2023} also uses text conditioning. Control-A-Video~\cite{ControlAVideo2023} first extracts Canny edges from the input video, then synthesizes the output video conditioned jointly on the edges and text.

For baselines without a video as input, we use the strategy contributed by a public pull request\footnote{https://github.com/guoyww/AnimateDiff/pull/8} to make AnimateDiff~\cite{AnimateDiff2023} take additional image conditioning. Specifically, at test time, latent features are first extracted from the condition image with the pre-trained SD VAE encoder~\cite{StableDiffusion2022}, which are then merged with the original per-frame Gaussian noises through linear blending. The diffusion process is later conditioned on a text prompt too. VideoCrafter1~\cite{VideoCrafter2023} provides both a text-to-video and an image-to-video model. We use the latter for a closer comparison setting.

\begin{figure}
    \centering
    \includegraphics[width=\linewidth]{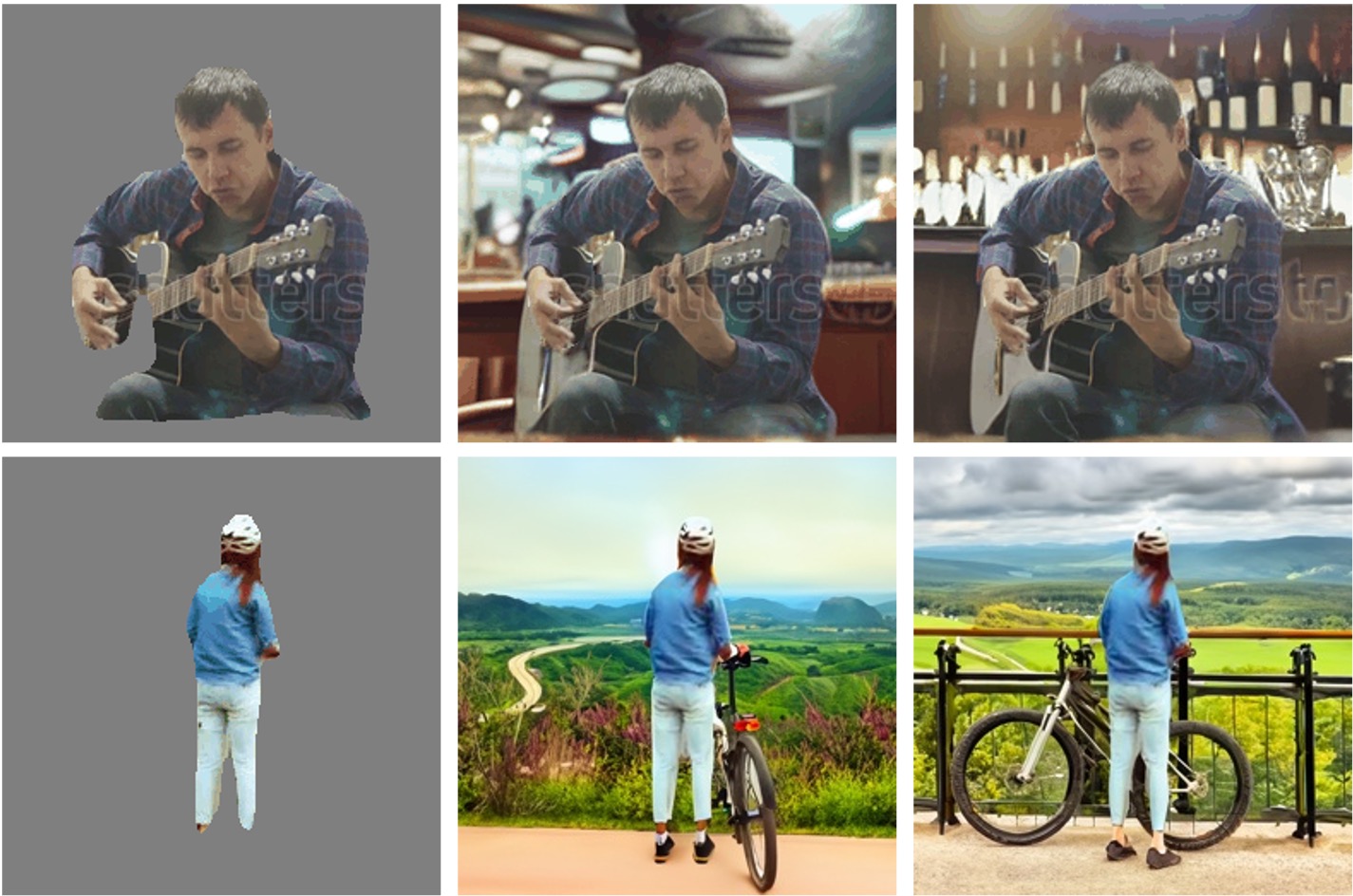}
    \vspace{-16px}
    \caption{Our method is robust to inaccurate masks. We show two examples from HiC+, each with its foreground segmentation followed by two generated outputs with different condition frames. We only show one frame and do not show the condition frame due to space limit. Please see supplement for full examples in videos.}
    \vspace{-4mm}
    \label{fig:inaccurate_mask}
\end{figure}

The qualitative comparison on two examples from the DAVIS~\cite{DAVIS2017} dataset is shown in \cref{fig:exp_comparison}. Our method generates temporally coherent videos that follow the foreground motion with highly realistic details, \eg falling snow and snow on the car windshield, while strictly follows the guidance and constraints given by the condition frame. Baseline methods in the first category generally inherit the structure present in the input video, \eg road direction, horse, etc., and hence they completely fail when fine-grained edits are desired, \eg horse changes to motorcycle in the second case. Methods in the second category generate unconstrained motion due to lack of guidance (VideoCrafter1 in the second example generates backward motion, which is more evident in the supplementary video).

\subsection{Additional Results and Analysis}
\label{subsec:addt_results}
\parahead{General video inpainting}
Interestingly, once trained, certain general video inpainting capability emerges from our model. We perform preliminary experiments by manually creating a mask sequence, and pass those with the foreground sequence as the input to our model, and we disable the condition signal by setting it to 0. Two cases are shown in \cref{fig:inpainting_app}, where our model is able to inpaint the missing regions, despite not explicitly trained so. This may suggest that our model learns to approximate the underlying data distribution to a certain degree, possibly benefiting from the random condition dropping during training (\cref{subsec:training}). We find similar results with general video outpainting, which we show in supplement.

\parahead{Robust to inaccurate masks}
As stated in \cref{subsec:training}, masks created or extracted in practice are often imperfect, being either incomplete or excessive. Here we show that our model trained in our designed procedure is robust to imperfect masks. In~\cref{fig:inaccurate_mask}, we showcase two examples of this. Despite a large region of the guitar (top) and both feet (bottom) missing, our model is able to hallucinate them in a reasonable way by considering the global context.

\parahead{Runtime}
Generating one video on an NVIDIA A100 GPU takes about 8.5 seconds, thus enables much faster idea iteration compared to traditional workflows.
\section{Conclusion}
We present \name, a video diffusion-based model that generates videos with coherent and vivid foreground-background interactions, given an input foreground segmentation sequence and a condition frame describing the background. Our model synthesizes highly realistic details such as moving or interacting objects and shadows. The generated videos also exhibit consistent camera scales and lighting effects. We believe our work contributes a useful tool for the movie and visual effects community, as well as for the general public to realize novel ideas of situating an acting subject in diverse scenes, in a simple and efficient way that is not previously possible. 
\section{Acknowledgment}
We thank the authors of~\cite{PutPeople2023} for compiling and processing the dataset HiC+, especially Sumith Kulal for the code and instructions on accessing the data. We also thank Jiahui (Gabriel) Huang from Adobe Research for helping set up the Adobe Firefly GenFill API.

{
    \small
    \bibliographystyle{ieeenat_fullname}
    \bibliography{main}

\begin{thebibliography}{47}
\providecommand{\natexlab}[1]{#1}
\providecommand{\url}[1]{\texttt{#1}}
\expandafter\ifx\csname urlstyle\endcsname\relax
  \providecommand{\doi}[1]{doi: #1}\else
  \providecommand{\doi}{doi: \begingroup \urlstyle{rm}\Url}\fi

\bibitem[Vir()]{VirtualProduction}
\emph{Virtual Production}.
\newblock \url{https://en.wikipedia.org/wiki/On-set_virtual_production}.

\bibitem[{Adobe}(2023{\natexlab{a}})]{adobe_firefly}
{Adobe}.
\newblock \emph{Firefly}.
\newblock 2023{\natexlab{a}}.
\newblock \url{https://www.adobe.com/sensei/generative-ai/firefly.html}.

\bibitem[{Adobe}(2023{\natexlab{b}})]{adobe_ps}
{Adobe}.
\newblock \emph{Photoshop, version}.
\newblock 2023{\natexlab{b}}.
\newblock \url{https://www.adobe.com/products/photoshop.html}.

\bibitem[Bain et~al.(2021)Bain, Nagrani, Varol, and Zisserman]{Bain21}
Max Bain, Arsha Nagrani, G{\"u}l Varol, and Andrew Zisserman.
\newblock Frozen in time: A joint video and image encoder for end-to-end retrieval.
\newblock In \emph{ICCV}, 2021.

\bibitem[Bar-Tal et~al.(2022)Bar-Tal, Ofri-Amar, Fridman, Kasten, and Dekel]{Text2LIVE2022}
Omer Bar-Tal, Dolev Ofri-Amar, Rafail Fridman, Yoni Kasten, and Tali Dekel.
\newblock Text2live: Text-driven layered image and video editing.
\newblock In \emph{ECCV}, 2022.

\bibitem[Blattmann et~al.(2023)Blattmann, Rombach, Ling, Dockhorn, Kim, Fidler, and Kreis]{AlignLatents2023}
Andreas Blattmann, Robin Rombach, Huan Ling, Tim Dockhorn, Seung~Wook Kim, Sanja Fidler, and Karsten Kreis.
\newblock Align your latents: High-resolution video synthesis with latent diffusion models.
\newblock In \emph{CVPR}, 2023.

\bibitem[Brooks and Efros(2022)]{brooks2022hallucinating}
Tim Brooks and Alexei~A Efros.
\newblock Hallucinating pose-compatible scenes.
\newblock In \emph{ECCV}, 2022.

\bibitem[Ceylan et~al.(2023)Ceylan, Huang, and Mitra]{Pix2Video2023}
Duygu Ceylan, Chun-Hao~P Huang, and Niloy~J Mitra.
\newblock Pix2video: Video editing using image diffusion.
\newblock In \emph{ICCV}, 2023.

\bibitem[Chen et~al.(2023{\natexlab{a}})Chen, Xia, He, Zhang, Cun, Yang, Xing, Liu, Chen, Wang, et~al.]{VideoCrafter2023}
Haoxin Chen, Menghan Xia, Yingqing He, Yong Zhang, Xiaodong Cun, Shaoshu Yang, Jinbo Xing, Yaofang Liu, Qifeng Chen, Xintao Wang, et~al.
\newblock Videocrafter1: Open diffusion models for high-quality video generation.
\newblock \emph{arXiv preprint arXiv:2310.19512}, 2023{\natexlab{a}}.

\bibitem[Chen et~al.(2023{\natexlab{b}})Chen, Wu, Xie, Wu, Li, Xia, Xiao, and Lin]{ControlAVideo2023}
Weifeng Chen, Jie Wu, Pan Xie, Hefeng Wu, Jiashi Li, Xin Xia, Xuefeng Xiao, and Liang Lin.
\newblock Control-a-video: Controllable text-to-video generation with diffusion models.
\newblock \emph{arXiv preprint arXiv:2305.13840}, 2023{\natexlab{b}}.

\bibitem[Clark et~al.(2019)Clark, Donahue, and Simonyan]{GANVideoGen2019}
Aidan Clark, Jeff Donahue, and Karen Simonyan.
\newblock Adversarial video generation on complex datasets.
\newblock \emph{arXiv preprint arXiv:1907.06571}, 2019.

\bibitem[Diba et~al.(2020)Diba, Fayyaz, Sharma, Paluri, Gall, Stiefelhagen, and Van~Gool]{hvu}
Ali Diba, Mohsen Fayyaz, Vivek Sharma, Manohar Paluri, J{\"u}rgen Gall, Rainer Stiefelhagen, and Luc Van~Gool.
\newblock Large scale holistic video understanding.
\newblock In \emph{ECCV}, 2020.

\bibitem[Esser et~al.(2023)Esser, Chiu, Atighehchian, Granskog, and Germanidis]{Gen12023}
Patrick Esser, Johnathan Chiu, Parmida Atighehchian, Jonathan Granskog, and Anastasis Germanidis.
\newblock Structure and content-guided video synthesis with diffusion models.
\newblock In \emph{ICCV}, 2023.

\bibitem[Fan et~al.(2023)Fan, Guo, Gong, Wang, Ge, Jiang, Luo, and Zhan]{M3DDM2023}
Fanda Fan, Chaoxu Guo, Litong Gong, Biao Wang, Tiezheng Ge, Yuning Jiang, Chunjie Luo, and Jianfeng Zhan.
\newblock Hierarchical masked 3d diffusion model for video outpainting.
\newblock In \emph{ACM MM}, 2023.

\bibitem[Ge et~al.(2022)Ge, Hayes, Yang, Yin, Pang, Jacobs, Huang, and Parikh]{TATS2022}
Songwei Ge, Thomas Hayes, Harry Yang, Xi Yin, Guan Pang, David Jacobs, Jia-Bin Huang, and Devi Parikh.
\newblock Long video generation with time-agnostic vqgan and time-sensitive transformer.
\newblock In \emph{ECCV}, 2022.

\bibitem[Geyer et~al.(2023)Geyer, Bar-Tal, Bagon, and Dekel]{TokenFlow2023}
Michal Geyer, Omer Bar-Tal, Shai Bagon, and Tali Dekel.
\newblock Tokenflow: Consistent diffusion features for consistent video editing.
\newblock \emph{arXiv preprint arXiv:2307.10373}, 2023.

\bibitem[Guo et~al.(2023)Guo, Yang, Rao, Wang, Qiao, Lin, and Dai]{AnimateDiff2023}
Yuwei Guo, Ceyuan Yang, Anyi Rao, Yaohui Wang, Yu Qiao, Dahua Lin, and Bo Dai.
\newblock Animatediff: Animate your personalized text-to-image diffusion models without specific tuning.
\newblock \emph{arXiv preprint arXiv:2307.04725}, 2023.

\bibitem[He et~al.(2017)He, Gkioxari, Doll{\'a}r, and Girshick]{he2017mask}
Kaiming He, Georgia Gkioxari, Piotr Doll{\'a}r, and Ross Girshick.
\newblock Mask r-cnn.
\newblock In \emph{ICCV}, 2017.

\bibitem[He et~al.(2023)He, Yang, Zhang, Shan, and Chen]{LVDM2022}
Yingqing He, Tianyu Yang, Yong Zhang, Ying Shan, and Qifeng Chen.
\newblock Latent video diffusion models for high-fidelity long video generation.
\newblock \emph{arXiv preprint arXiv:2211.13221}, 2023.

\bibitem[Ho et~al.(2020)Ho, Jain, and Abbeel]{DDPM2020}
Jonathan Ho, Ajay Jain, and Pieter Abbeel.
\newblock Denoising diffusion probabilistic models.
\newblock In \emph{NeurIPS}, 2020.

\bibitem[Ho et~al.(2022)Ho, Chan, Saharia, Whang, Gao, Gritsenko, Kingma, Poole, Norouzi, Fleet, et~al.]{ImagenVideo2022}
Jonathan Ho, William Chan, Chitwan Saharia, Jay Whang, Ruiqi Gao, Alexey Gritsenko, Diederik~P Kingma, Ben Poole, Mohammad Norouzi, David~J Fleet, et~al.
\newblock Imagen video: High definition video generation with diffusion models.
\newblock \emph{arXiv preprint arXiv:2210.02303}, 2022.

\bibitem[Hu and Xu(2023)]{VideoControlNet2023}
Zhihao Hu and Dong Xu.
\newblock Videocontrolnet: A motion-guided video-to-video translation framework by using diffusion model with controlnet.
\newblock \emph{arXiv preprint arXiv:2307.14073}, 2023.

\bibitem[Kasten et~al.(2021)Kasten, Ofri, Wang, and Dekel]{NLA2021}
Yoni Kasten, Dolev Ofri, Oliver Wang, and Tali Dekel.
\newblock Layered neural atlases for consistent video editing.
\newblock \emph{ACM TOG}, 40\penalty0 (6), 2021.

\bibitem[Khachatryan et~al.(2023)Khachatryan, Movsisyan, Tadevosyan, Henschel, Wang, Navasardyan, and Shi]{T2VZero2023}
Levon Khachatryan, Andranik Movsisyan, Vahram Tadevosyan, Roberto Henschel, Zhangyang Wang, Shant Navasardyan, and Humphrey Shi.
\newblock Text2video-zero: Text-to-image diffusion models are zero-shot video generators.
\newblock \emph{arXiv preprint arXiv:2303.13439}, 2023.

\bibitem[Kingma and Welling(2013)]{kingma2013auto}
Diederik~P Kingma and Max Welling.
\newblock Auto-encoding variational bayes.
\newblock \emph{arXiv preprint arXiv:1312.6114}, 2013.

\bibitem[Kulal et~al.(2023)Kulal, Brooks, Aiken, Wu, Yang, Lu, Efros, and Singh]{PutPeople2023}
Sumith Kulal, Tim Brooks, Alex Aiken, Jiajun Wu, Jimei Yang, Jingwan Lu, Alexei~A Efros, and Krishna~Kumar Singh.
\newblock Putting people in their place: Affordance-aware human insertion into scenes.
\newblock In \emph{CVPR}, 2023.

\bibitem[Loshchilov and Hutter(2017)]{AdamW2017}
Ilya Loshchilov and Frank Hutter.
\newblock Decoupled weight decay regularization.
\newblock \emph{arXiv preprint arXiv:1711.05101}, 2017.

\bibitem[Monfort et~al.(2019)Monfort, Andonian, Zhou, Ramakrishnan, Bargal, Yan, Brown, Fan, Gutfreund, Vondrick, et~al.]{monfort2019moments}
Mathew Monfort, Alex Andonian, Bolei Zhou, Kandan Ramakrishnan, Sarah~Adel Bargal, Tom Yan, Lisa Brown, Quanfu Fan, Dan Gutfreund, Carl Vondrick, et~al.
\newblock Moments in time dataset: one million videos for event understanding.
\newblock \emph{IEEE TPAMI}, 42\penalty0 (2), 2019.

\bibitem[{OpenAI}(2023)]{ChatGPT}
{OpenAI}.
\newblock \emph{ChatGPT}.
\newblock 2023.
\newblock \url{https://chat.openai.com/}.

\bibitem[Pont-Tuset et~al.(2017)Pont-Tuset, Perazzi, Caelles, Arbel{\'a}ez, Sorkine-Hornung, and Van~Gool]{DAVIS2017}
Jordi Pont-Tuset, Federico Perazzi, Sergi Caelles, Pablo Arbel{\'a}ez, Alex Sorkine-Hornung, and Luc Van~Gool.
\newblock The 2017 davis challenge on video object segmentation.
\newblock \emph{arXiv preprint arXiv:1704.00675}, 2017.

\bibitem[Radford et~al.(2021)Radford, Kim, Hallacy, Ramesh, Goh, Agarwal, Sastry, Askell, Mishkin, Clark, et~al.]{CLIP2021}
Alec Radford, Jong~Wook Kim, Chris Hallacy, Aditya Ramesh, Gabriel Goh, Sandhini Agarwal, Girish Sastry, Amanda Askell, Pamela Mishkin, Jack Clark, et~al.
\newblock Learning transferable visual models from natural language supervision.
\newblock In \emph{ICML}, 2021.

\bibitem[Ramesh et~al.(2022)Ramesh, Dhariwal, Nichol, Chu, and Chen]{DALLE2}
Aditya Ramesh, Prafulla Dhariwal, Alex Nichol, Casey Chu, and Mark Chen.
\newblock Hierarchical text-conditional image generation with clip latents.
\newblock \emph{arXiv preprint arXiv:2204.06125}, 2022.

\bibitem[Ranftl et~al.(2020)Ranftl, Lasinger, Hafner, Schindler, and Koltun]{MiDaS2020}
Ren{\'e} Ranftl, Katrin Lasinger, David Hafner, Konrad Schindler, and Vladlen Koltun.
\newblock Towards robust monocular depth estimation: Mixing datasets for zero-shot cross-dataset transfer.
\newblock \emph{IEEE TPAMI}, 44\penalty0 (3), 2020.

\bibitem[Rombach et~al.(2022)Rombach, Blattmann, Lorenz, Esser, and Ommer]{StableDiffusion2022}
Robin Rombach, Andreas Blattmann, Dominik Lorenz, Patrick Esser, and Bj{\"o}rn Ommer.
\newblock High-resolution image synthesis with latent diffusion models.
\newblock In \emph{CVPR}, 2022.

\bibitem[Ronneberger et~al.(2015)Ronneberger, Fischer, and Brox]{U-Net}
Olaf Ronneberger, Philipp Fischer, and Thomas Brox.
\newblock U-net: Convolutional networks for biomedical image segmentation.
\newblock In \emph{MICCAI}, 2015.

\bibitem[Saharia et~al.(2022)Saharia, Chan, Chang, Lee, Ho, Salimans, Fleet, and Norouzi]{Palette2022}
Chitwan Saharia, William Chan, Huiwen Chang, Chris Lee, Jonathan Ho, Tim Salimans, David Fleet, and Mohammad Norouzi.
\newblock Palette: Image-to-image diffusion models.
\newblock In \emph{ACM SIGGRAPH}, 2022.

\bibitem[Sigurdsson et~al.(2016)Sigurdsson, Varol, Wang, Farhadi, Laptev, and Gupta]{sigurdsson2016hollywood}
Gunnar~A Sigurdsson, G{\"u}l Varol, Xiaolong Wang, Ali Farhadi, Ivan Laptev, and Abhinav Gupta.
\newblock Hollywood in homes: Crowdsourcing data collection for activity understanding.
\newblock In \emph{ECCV}, 2016.

\bibitem[Tulyakov et~al.(2018)Tulyakov, Liu, Yang, and Kautz]{MOCOGAN2018}
Sergey Tulyakov, Ming-Yu Liu, Xiaodong Yang, and Jan Kautz.
\newblock Mocogan: Decomposing motion and content for video generation.
\newblock In \emph{CVPR}, 2018.

\bibitem[Villegas et~al.(2022)Villegas, Babaeizadeh, Kindermans, Moraldo, Zhang, Saffar, Castro, Kunze, and Erhan]{Phenaki2022}
Ruben Villegas, Mohammad Babaeizadeh, Pieter-Jan Kindermans, Hernan Moraldo, Han Zhang, Mohammad~Taghi Saffar, Santiago Castro, Julius Kunze, and Dumitru Erhan.
\newblock Phenaki: Variable length video generation from open domain textual description.
\newblock \emph{arXiv preprint arXiv:2210.02399}, 2022.

\bibitem[Vondrick et~al.(2016)Vondrick, Pirsiavash, and Torralba]{SceneDynam2016}
Carl Vondrick, Hamed Pirsiavash, and Antonio Torralba.
\newblock Generating videos with scene dynamics.
\newblock In \emph{NeurIPS}, 2016.

\bibitem[Wu et~al.(2023)Wu, Ge, Wang, Lei, Gu, Shi, Hsu, Shan, Qie, and Shou]{TuneAVideo2023}
Jay~Zhangjie Wu, Yixiao Ge, Xintao Wang, Stan~Weixian Lei, Yuchao Gu, Yufei Shi, Wynne Hsu, Ying Shan, Xiaohu Qie, and Mike~Zheng Shou.
\newblock Tune-a-video: One-shot tuning of image diffusion models for text-to-video generation.
\newblock In \emph{ICCV}, 2023.

\bibitem[Xie et~al.(2023)Xie, Zhang, Lin, Hinz, and Zhang]{SmartBrush2023}
Shaoan Xie, Zhifei Zhang, Zhe Lin, Tobias Hinz, and Kun Zhang.
\newblock Smartbrush: Text and shape guided object inpainting with diffusion model.
\newblock In \emph{CVPR}, 2023.

\bibitem[Yang et~al.(2023)Yang, Gu, Zhang, Zhang, Chen, Sun, Chen, and Wen]{PaintbyExample2023}
Binxin Yang, Shuyang Gu, Bo Zhang, Ting Zhang, Xuejin Chen, Xiaoyan Sun, Dong Chen, and Fang Wen.
\newblock Paint by example: Exemplar-based image editing with diffusion models.
\newblock In \emph{CVPR}, 2023.

\bibitem[Yu et~al.(2023{\natexlab{a}})Yu, Cheng, Sohn, Lezama, Zhang, Chang, Hauptmann, Yang, Hao, Essa, et~al.]{MAGVIT2023}
Lijun Yu, Yong Cheng, Kihyuk Sohn, Jos{\'e} Lezama, Han Zhang, Huiwen Chang, Alexander~G Hauptmann, Ming-Hsuan Yang, Yuan Hao, Irfan Essa, et~al.
\newblock Magvit: Masked generative video transformer.
\newblock In \emph{CVPR}, 2023{\natexlab{a}}.

\bibitem[Yu et~al.(2023{\natexlab{b}})Yu, Sohn, Kim, and Shin]{PVDM2023}
Sihyun Yu, Kihyuk Sohn, Subin Kim, and Jinwoo Shin.
\newblock Video probabilistic diffusion models in projected latent space.
\newblock In \emph{CVPR}, 2023{\natexlab{b}}.

\bibitem[Zhang et~al.(2023)Zhang, Rao, and Agrawala]{ControlNet2023}
Lvmin Zhang, Anyi Rao, and Maneesh Agrawala.
\newblock Adding conditional control to text-to-image diffusion models.
\newblock In \emph{CVPR}, 2023.

\bibitem[Zhou et~al.(2023)Zhou, Li, Chan, and Loy]{ProPainter2023}
Shangchen Zhou, Chongyi Li, Kelvin~CK Chan, and Chen~Change Loy.
\newblock Propainter: Improving propagation and transformer for video inpainting.
\newblock In \emph{ICCV}, 2023.

\end{thebibliography}
}

\newpage
\clearpage
\appendix

\twocolumn[{
\renewcommand\twocolumn[1][]{#1}
\maketitle
\begin{center}
    \centering
    \includegraphics[width=\linewidth]{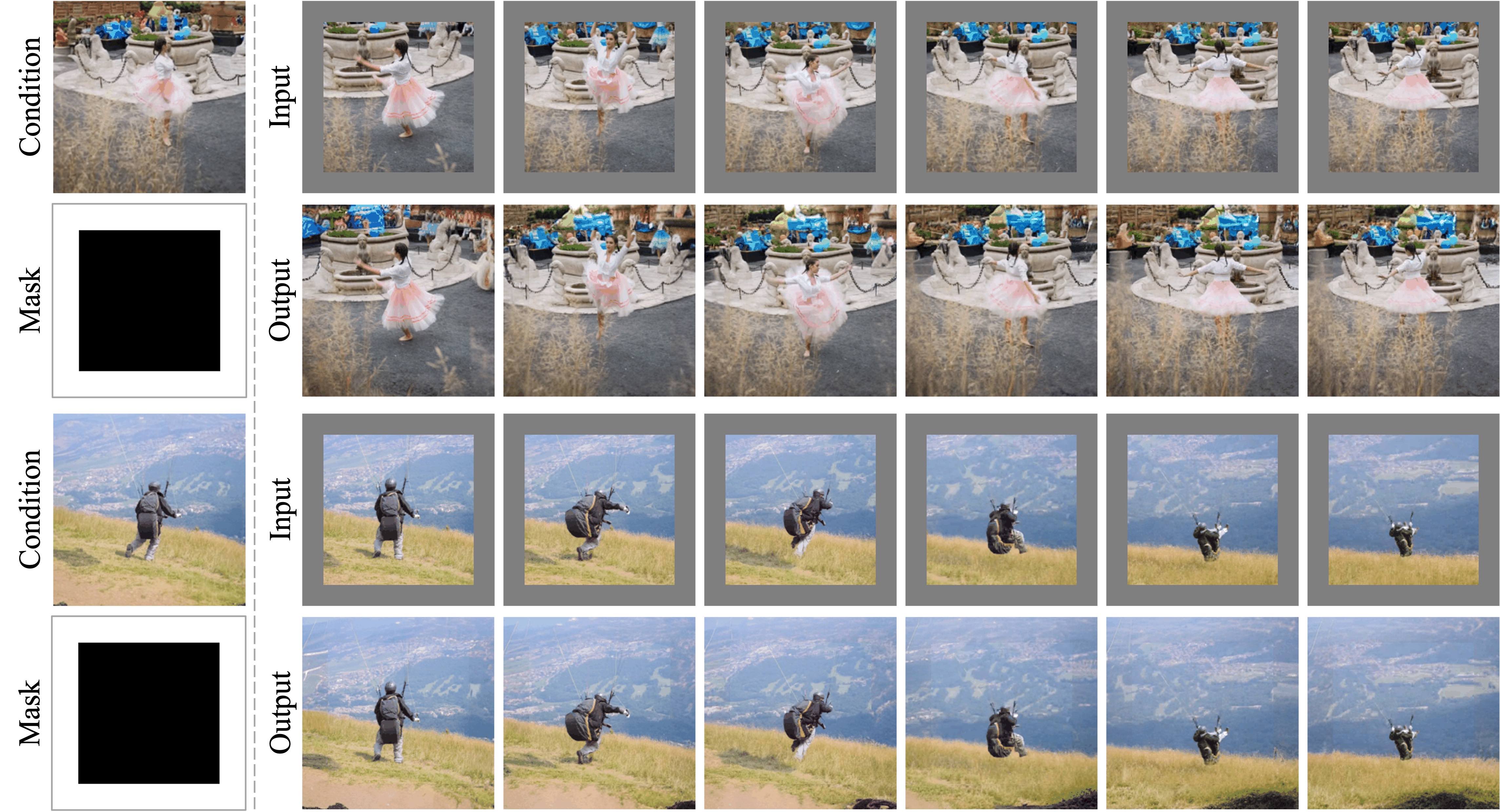}
    \captionof{figure}{\textbf{Zero-shot video outpainting with our model.} We show two examples from the DAVIS dataset.}
    \label{fig:outpaint_app}
\end{center}
}
]

In this supplement, we first provide more examples of the general video outpainting application and of that our model is robust to inaccurate masks in~\cref{supp_sec:full_vis}, following Sec.~4.3 of the main manuscript. We then describe essential processing steps for training and evaluation data in~\cref{supp_sec:data_process}. We show failure cases and discuss limitations of our model in~\cref{supp_sec:limitations}. Lastly, we conclude by discussing the ethical impact of this work in ~\cref{supp_sec:ethics}. 

We strongly encourage the reader to check our project webpage, where we show extensive video results on video background generation with diverse generated contents and camera motions, and under various conditioning scenarios. It also contains the video version of the comparison with baselines. 

\section{Full Examples for Sec.~4.3 of the Main Paper}
\label{supp_sec:full_vis}
\parahead{General video outpainting}
As stated in Sec.~4.3 of the main manuscript, preliminary capability on general video inpainting / outpainting emerges from our model once trained. In Fig.~5 of the main paper, we show results on general video inpainting. Here we present the results for applying the same model on general video outpainting in~\cref{fig:outpaint_app}. 
We resize the original sequence of frames by 0.75, and pad them with gray boundaries. Associated masks are also created to indicate the missing regions. We then randomly select one frame and use Adobe Firefly~\cite{adobe_firefly} to outpaint it, with which as condition we outpaint the entire sequence.
Along with the general video inpainting results shown in Fig.~5 of the main paper, these suggest that our model learns to approximate the underlying data distribution to a certain degree.

\begin{figure*}
    \centering
    \includegraphics[width=\linewidth]{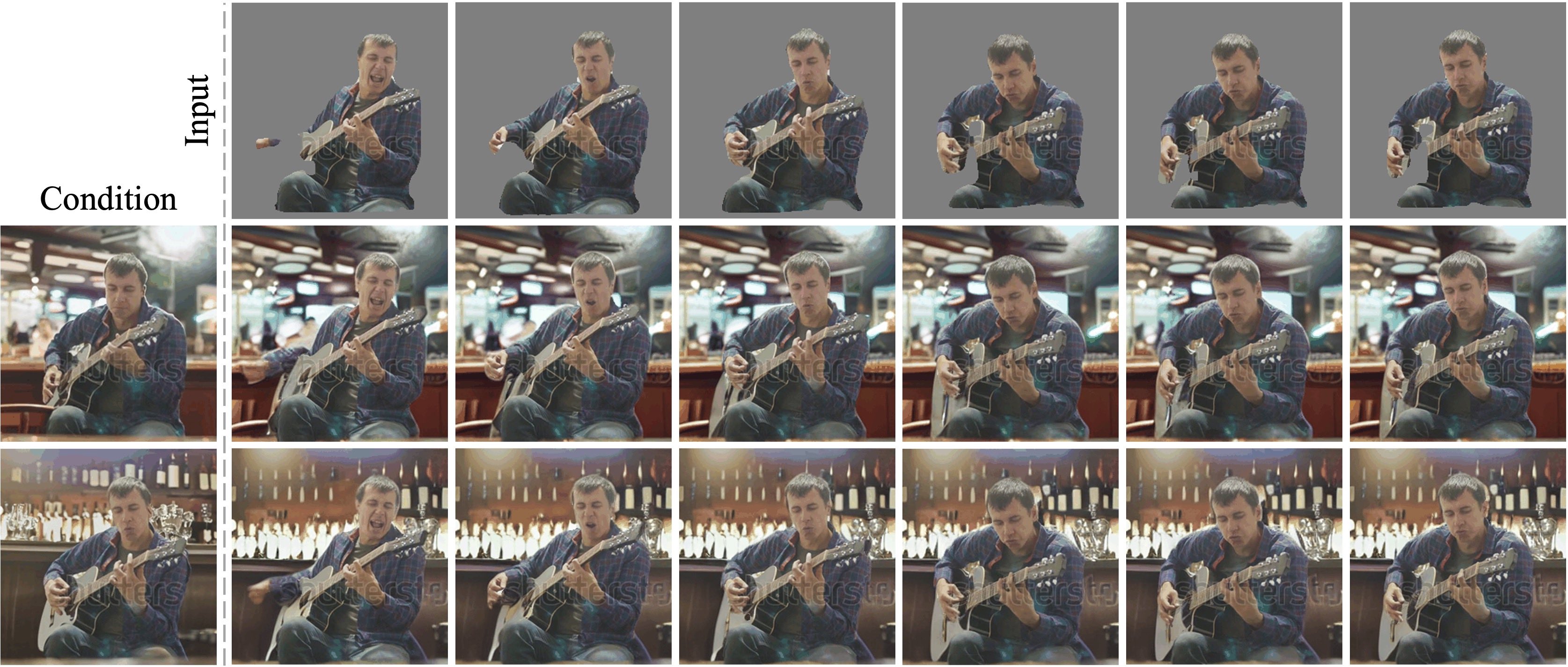}
    \caption{Our model is robust to inaccurate masks. Here we show the full version of the first example in Fig.~6 of the main manuscript.}
    \label{fig:full_inaccurate_mask}
\end{figure*}

\begin{figure}
    \centering
    \includegraphics[width=\linewidth]{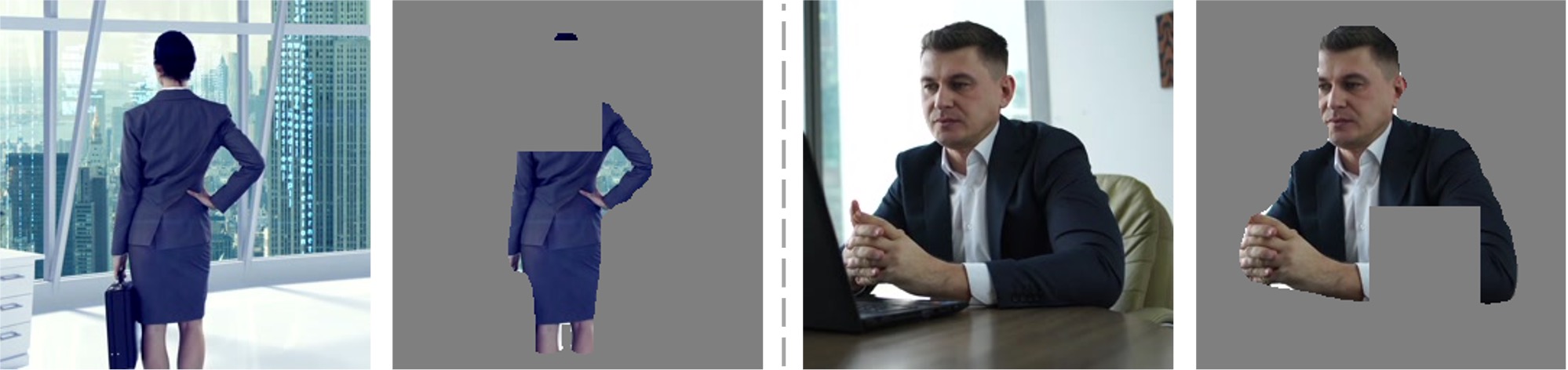}
    \caption{
    \textbf{Data augmentation}. We apply random cut-outs to the person segmentation during training. Here we show two examples of cut-outs with their corresponding original frames.
    }
    \label{fig:augmentation}
\end{figure}

\parahead{Robust to inaccurate masks}
We show the full version of the first example in Fig.~6 of the main manuscript in~\cref{fig:full_inaccurate_mask}. Our model fills in realistic details of the guitar and the man's hand, which are missing from the segmentation due to inaccurate masks. 

\section{Data Processing}
\label{supp_sec:data_process}
\parahead{Training}
In Sec.~3.4 of the main manuscript, we described our data processing and augmentation strategies for training data. Specifically, to deal with incomplete segmentation, we apply random rectangular cut-outs to the segmentation and masks. We show two examples in~\cref{fig:augmentation}.

\parahead{Evaluation}
As mentioned in Sec.~1 of the main manuscript, at test time, the composited foreground-background frames used as condition can be created with various methods, such as photo editing tools (\eg Adobe Photoshop~\cite{adobe_ps}) or automated image outpainting methods (\eg Dall-E~\cite{DALLE2}). In our experiments, we adopt ChatGPT 4~\cite{ChatGPT} and Adobe Firefly~\cite{adobe_firefly} to automatically synthesize these composited frames for use at test time. Specifically, we first sample the 0-th, 8-th, and 15-th frames from the input video, and ask ChatGPT the following question ``Based on these three images, can you give me a description as prompt, less than 10 words, about one contextual scenario we can put this human in?". We then use Firefly to synthesize the outpainted frames, given the foreground segmentation and the text prompt. We use the ``gpt4-vision-preview" version of ChatGPT 4~\footnote{https://platform.openai.com/docs/overview}, and the official Firefly GenFill~\footnote{https://firefly.adobe.com/generate/inpaint}.

\begin{figure}
    \centering
    \includegraphics[width=0.9\linewidth]{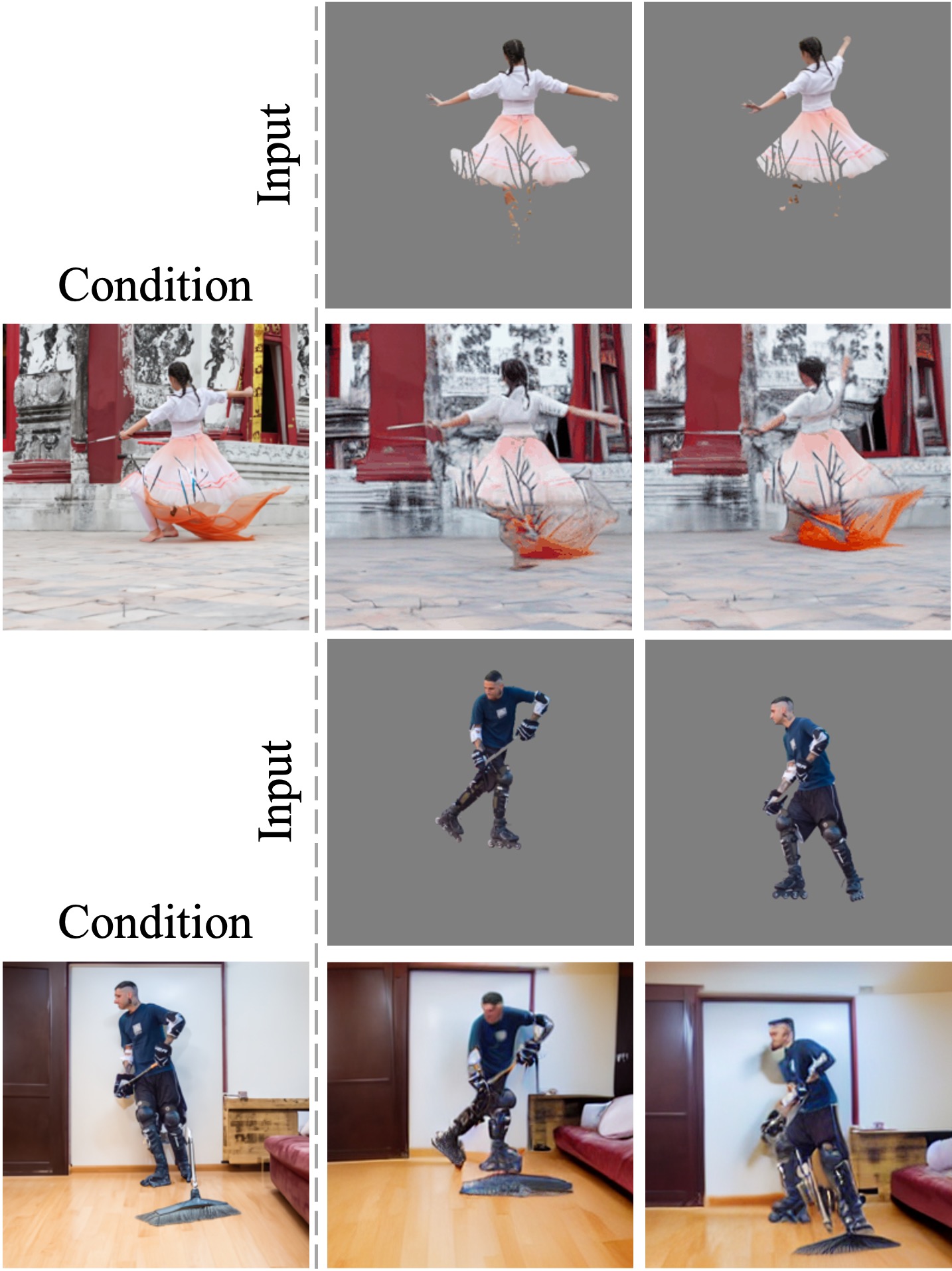}
    \caption{
    \textbf{Failure cases}. Foreground sequences from DAVIS.
    }
    \label{fig:failure_cases}
\end{figure}

\section{Limitations}
\label{supp_sec:limitations}
We show two failure cases of our method in~\cref{fig:failure_cases}. In the first example, the grass-like texture on the dress is excluded from the segmentation, hence the model mistakenly perceives it to be an independent object growing outside the dress. While in the second example, the Firefly-inpainted frame has the broomstick facing towards the wrong direction. Although the model tries to generate something reasonable, it fails to correct this mistake to produce a coherent video.
Despite certain fault tolerance from our model, providing proper input segmentation and condition signal helps ensure high-quality generation results.

\section{Data Ethics}
\label{supp_sec:ethics}
The HiC+ dataset~\cite{PutPeople2023} includes public datasets HVU~\cite{hvu}, Charades~\cite{sigurdsson2016hollywood}, Moments~\cite{monfort2019moments}, and WebVid10M~\cite{Bain21}. Readers are encouraged to refer to Section A.1 in the supplementary material of~\cite{brooks2022hallucinating} for the license of these datasets. 
Moreover, same as~\cite{PutPeople2023}, more internal videos are incorporated during training and evaluation. We have conducted the necessary legal review process, and can provide more details of this data upon request.

We present a method that can synthesize highly realistic videos that place human and non-human subjects into diverse background scenes. While our work has implications and values for a wide range of fields, we note that our model can be misused to generate malicious content. Similar to Firefly GenFill~\cite{adobe_firefly} and Gen1~\cite{Gen12023}, the generated content can be watermarked to prevent misuse. Moreover, our model inherits the demographic biases presented in the training data. We
make our best effort to demonstrate impartiality in all the results shown.


\end{document}